\documentclass[10pt,twocolumn,letterpaper]{article}

\usepackage[pagenumbers]{cvpr} 

\usepackage{colortbl} 
\usepackage[dvipsnames]{xcolor}
\usepackage{arydshln} %dashed horizontal line in table
\usepackage{amssymb}% http://ctan.org/pkg/amssymb
\usepackage{pifont}% http://ctan.org/pkg/pifont
\newcommand{\cmark}{\ding{51}}
\newcommand{\xmark}{\ding{55}}
\usepackage{comment}
\usepackage{cuted}
\definecolor{cvprblue}{rgb}{0.21,0.49,0.74}
\usepackage[pagebackref,breaklinks,colorlinks,allcolors=cvprblue]{hyperref}

\title{\textsc{JASPR}: Joint Spatial Representation learning of histology and spatial genomics for improved virtual genomic screening and clinical prognostication}

\author{
\makebox[\textwidth][c]{%
Marija Pizurica$^{1,2}$\hspace{2.0em}%
Eric Zimmermann$^{1}$\hspace{2.0em}%
Neil Tenenholtz$^{1}$\hspace{2.0em}%
James Hall$^{1}$%
}\\[0.2em]
\makebox[\textwidth][c]{%
Olivier Gevaert$^{2}$\hspace{2.0em}%
Ava P. Amini$^{1}$\hspace{2.0em}%
Lorin Crawford$^{1}$\hspace{2.0em}%
Kristen A. Severson$^{1}$%
}\\[0.5em]
\makebox[\textwidth][c]{$^{1}$Microsoft Research, Cambridge, MA, United States \qquad
$^{2}$Stanford University, CA, United States}\\[0.2em]
\makebox[\textwidth][c]{{\small Corresponding author: \texttt{kseverson@microsoft.com}}}
}

\begin{document}
\maketitle

\begin{abstract}
    Recent studies have shown that spatial properties of tumors are critical for understanding disease biology and predicting patient outcomes. These spatial properties are increasingly uncovered through complementary modalities: spatial transcriptomics (ST) captures spatially-resolved molecular states, while hematoxylin and eosin-stained whole slide images (HE) reveal tissue morphology. While approaches are emerging to fuse these modalities, effective methods that learn not only joint representations but also incorporate spatial context across modalities are lacking. Here, we present \textsc{JASPR} (Joint Spatial Representation learning), a self-supervised deep learning framework that integrates HE images and ST data through a cross-modal reconstruction objective that incorporates spatial context within HE images and ST profiles. It employs shared modules to capture universal spatial properties across modalities, while modality-specific experts encode features unique to morphological and genomic data. We train and validate \textsc{JASPR} on breast cancer datasets, demonstrating that its learned joint representation substantially improves HE-based prediction of 9,248 genes and provides prognostic value for breast cancer outcomes. 
\end{abstract}

\section{Introduction}
\label{sec:intro}

Molecular profiling of tumor tissue is increasingly performed to characterize diagnostic and prognostic biomarkers, which in turn are used to guide treatment selection for cancer patients. Traditionally, this process has relied on bulk omics assays (e.g., whole-exome sequencing, RNA-seq, and targeted gene panels) that aggregate signals across specimens \cite{syed2020oncotype, slodkowska2009mammaprint, sestak2020prognostic, nielsen2010comparison}. Recently, studies analyzing spatial gene expression, called spatial transcriptomics (ST) data, have revealed that, by collapsing spatial molecular properties into averaged measurements, bulk profiling loses information on distinct microenvironments and their spatial arrangements, both of which can correlate strongly with prognosis and therapeutic response \cite{wang2024spatial, an2024spatial, jimenez2024spatial}. This has motivated a shift toward models that explicitly capture intratumoral heterogeneity and spatial context.

In parallel, hematoxylin and eosin-stained tissue slides (HE) remain central to the standard-of-care in pathology, and HE images are routinely acquired for cancer diagnosis and staging. The ongoing digitization of pathology workflows has transformed these images into a scalable data source for computational analysis. Deep learning applied to HE images has demonstrated success in clinically relevant tasks including tumor detection, grading, prognosis prediction, and treatment response \cite{jahan2025deep, liu2022deep, li2022deep, perera2024annotation}.

HE images and ST data provide complementary views of tissue organization: the former captures morphological structure and is readily available, while the latter resolves spatially organized molecular states but remains expensive and uncommon in standard workflows. Given the recent successes of foundation models in computational pathology, there is an open question in the field as to whether HE inputs can be used to create synthetic spatial transcriptomics profiles (i.e., virtual screenings) \cite{alsaafin2023learning, schmauch2020deep, pizurica2024digital, shulman2024ai, he2020integrating, nonchev2025deepspot, pang2021leveraging, jia2023thitogene}. Additionally, it is hypothesized that spatial transcriptomics data could serve as useful pretraining signal to improve HE representations \cite{chen2025visual, redekop2025spade, gindra2025large, xie2023spatially}. 

Effectively integrating these spatial, high-dimensional, and heterogeneous modalities remains a key challenge and is the focus of our work. We propose that successful integration requires modeling along three complementary axes: (1) spatial context within HE images, (2) spatial context within ST profiles, and (3) cross-modal joint representations. While existing methods address subsets of these axes, methods that comprehensively incorporate all three are lacking (Suppl. Table \ref*{tab:sota_overview}). 

To address this gap, we present \textsc{JASPR} (Joint SPatial Representation learning; Fig.~\ref{fig:architecture}), which leverages spatial context present within both HE and ST data and learns a joint representation informed by both modalities while preserving their spatial organization. We perform extensive ablative studies on different design decisions for multi-modal training. Based on the results, we then train \textsc{JASPR} models on breast cancer data, and we evaluate their learned representations on virtual ST prediction from HE images and on clinical prediction tasks. We show that both virtual screening and clinical tasks benefit from \textsc{JASPR} even when compared to models with contrastive pretraining but without spatial awareness.

\section{Background and related work}
HE images present unique computational challenges due to their enormous size, often exceeding 10k $\times$ 10k pixels at diagnostic resolution, and due to large morphological and molecular heterogeneity. To make these images tractable for deep learning models, the standard approach involves dividing the image into smaller tiles (e.g., 224 $\times$ 224) and then embedding these tiles with a foundation model \cite{jaume2024hest}. Two main modeling strategies are followed: tile-level models are trained to make predictions independently for each tile, and slide-level models are trained to aggregate information across tiles for the entire HE image. 

When training these models with additional molecular labels, data may be available at either bulk or spatial resolution. In particular, spatial transcriptomics technologies quantify gene expression while preserving spatial coordinates within tissue, enabling molecular measurements to be aligned with histology. ST assays can be sequencing-based such as 10x Genomics' Visium, and imaging-based such as 10x Genomics' Xenium. The latter provides higher spatial resolution but typically measures targeted gene panels. In Visium, RNA is collected at small tissue regions on the slide referred to as spots (often comprising multiple cells). These spot-level expression profiles can be matched to co-registered HE image tiles extracted from the same locations (Fig.~\ref{fig:architecture}, left panel). 

Here, we review existing approaches for predicting genomic profiles from HE images along three axes: modeling spatial context within HE images, within ST data, and learning joint cross-modal representations. Notably, current methods only partially address these dimensions. We then end this section with a review of spatial representation learning methods that inspire \textsc{JASPR}'s design.

\subsection{HE-based Virtual Genomic Screening}
The earliest approaches for HE-based virtual genomics predicted bulk molecular properties, such as DNA mutations and gene expression \cite{kather2020pan, pizurica2023whole, schaumberg2016h, chen2020classification, qu2021genetic, wang2021predicting}, from individual histology tiles. 
More recent work has used slide-based methods, aggregating information across tiles \cite{alsaafin2023learning, schmauch2020deep, pizurica2024digital}. Although the slide-based approaches are better positioned to address some of the tumor heterogeneity, both approaches are limited by irreducible label noise due to the lack of precise correspondence between the two modalities. 
 
Spatial transcriptomics resolves the challenge of label correspondence and some works have proposed either tile-based modeling (i.e., spatially-independent models) \cite{shulman2024ai, he2020integrating} or region-based modeling \cite{nonchev2025deepspot, pang2021leveraging, jia2023thitogene}, which incorporates some HE context. 
Despite these advances, all the aforementioned methods share two key limitations: they do not explicitly model spatial relationships within the genomics data, and they lack joint representations that integrate both modalities during training.

\subsection{Joint Modeling of HE and Genomics}
Unlike the approaches that use HE images solely as model input, joint learning methods align both modalities in a shared representation space, for example through contrastive learning. One such work \cite{jaume2024transcriptomics} introduces shared representations between slide-level HE images and bulk gene expression. Alternative methods \cite{chen2025visual, redekop2025spade, gindra2025large, xie2023spatially} apply contrastive learning to spatial transcriptomics, but still operate at the tile level. Most recently, a cross-modal fusion and reconstruction approach \cite{kong2025spatia} has been proposed that combines slide-level modeling with transcriptomics while incorporating additional image context. However, this method uses microscopy images rather than HE images and still does not explicitly model spatial relationships within the genomics data.

\subsection{Spatial Representation Learning}
Spatial representation learning aims to learn feature representations that preserve spatial structure, such as local neighborhoods and long-range contextual relationships, rather than collapsing observations into a single global embedding. In natural images, a typical modern approach is masked image modeling (MIM), in which a model is trained to infer missing content from visible context. A widely used instance is the masked autoencoder framework (MAE) \cite{he2022masked}, which reconstructs masked pixels/embeddings with a lightweight decoder. While alternative families can also produce spatially meaningful features, including contrastive learning \cite{chen2020simple, he2020momentum}, teacher-student self-distillation \cite{grill2020bootstrap, caron2021emerging}, or generative objectives \cite{ren2023rejuvenating}, we adopt an MAE-based formulation because it provides a simple, reconstruction-driven objective that does not require negatives and extends naturally to multimodal spatial data. Masking and reconstructing can be applied within and across modalities, enabling an elegant mechanism to model histology images and spatial transcriptomics jointly by learning shared representations that leverage spatial context in both modalities.

\section{Methodology}
\begin{figure*}
    \centering
    \includegraphics[width=\linewidth]{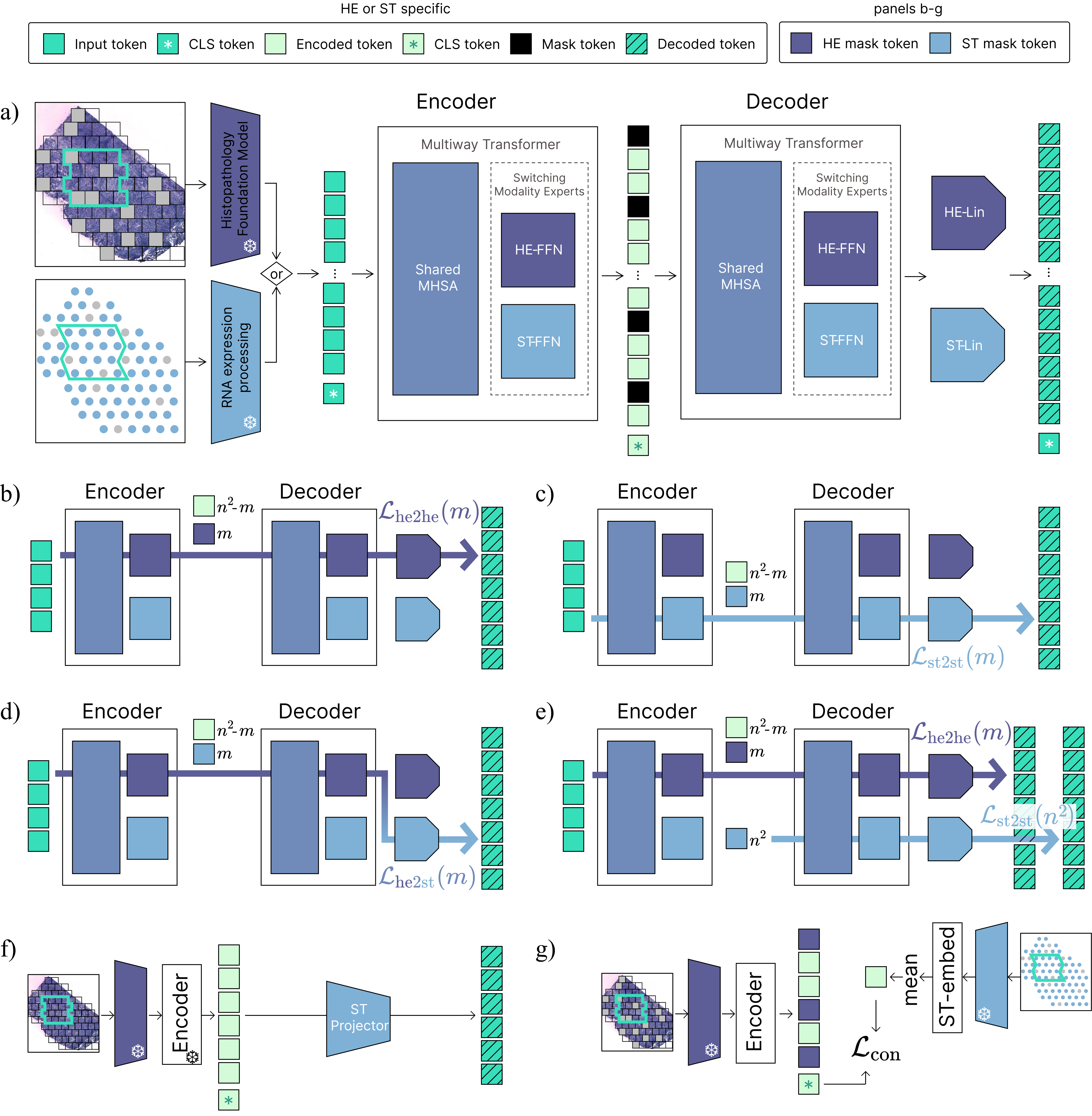}
    \caption{\textbf{Overview of \textsc{JASPR}}. \textbf{a)} The input to \textsc{JASPR} is a window of $n \times n$ tokens ($n=8$ throughout), representing HE embeddings extracted with Virchow2 ($d=1280$), or gene expression measurements from spatial transcriptomics data ($d=9248)$. (\textbf{b)-g)}) Learning objectives and input-output pairs for \textsc{JASPR}. \textbf{b)} HE2HE objective \textbf{c)} ST2ST objective \textbf{d)} HE2ST objective, note that the combination of HE2ST and HE2HE is the separate (-sep) setting \textbf{e)} Joint objective \textbf{f)} Virtual screening task \textbf{g)} Contrastive loss.} 
    \label{fig:architecture}
\end{figure*}

\textsc{JASPR} uses a transformer-based, asymmetric encoder-decoder model and MAE objective to achieve two key aims: (1) prediction of spatially resolved gene expression (i.e., virtual screening) and (2) enhanced representations of histopathology images for tasks with connection to molecular state (e.g., prognostication). As described above, these capabilities are motivated by the observation that spatial transcriptomics provides a rich description of molecular state but is infrequently available at inference time. We hypothesize that spatial masking is critical to representation learning and well-suited as a pretraining objective. Masking encourages the model to move beyond local morphological and gene-level signals, instead capturing higher-order tissue architecture and spatial dependencies. By focusing on the multimodal setting, \textsc{JASPR} learns shared latent representations, aligning histological patterns with underlying transcriptional signals. An overview of \textsc{JASPR} pretraining is presented in Fig.~\ref{fig:architecture} and described in detail in the following sections. 

\subsection{\textsc{JASPR} Model Architecture}
The most parsimonious \textsc{JASPR} design, which achieves the aims described above, is trained using masked modeling on an $n \times n$ grid of histopathology foundation model embeddings as input (abbreviated HE) and outputs a matched grid of $n \times n$ gene expression (abbreviated ST). However, given the success of unified masked data modeling objectives~\cite{wang2023image}, we also consider other tasks, namely embeddings to embeddings (HE2HE) and gene expression to gene expression (ST2ST). To enable joint modeling, \textsc{JASPR} (Fig.~\ref{fig:architecture}a) employs a multiway transformer design inspired by BEiT3 \cite{wang2023image, bao2022vlmo}. In the multiway transformer, the multi-head self-attention (MHSA) mechanism is shared across modalities to capture cross-modal alignments, while modality-specific feed-forward networks (HE-FFN and ST-FFN) allow the model to preserve unique features of each data type. The feed-forward networks are followed by linear projectors which map the encoded dimension back to the observed dimension space. The model also uses positional embeddings for HE and ST that are shared and kept fixed (i.e., not learned) to enforce a shared spatial reference. This allows cross-modal interactions to occur at corresponding spatial locations, and to capture complementary information tied to identical spatial coordinates. As in many transformer architectures, we include a learnable CLS token, which is modality-dependent, to aggregate global information across the input window and provide a compact representation for downstream tasks. 

We additionally consider the optional inclusion of an ST projector, which maps unmasked encoded HE tokens to observed gene expression (virtual screening task, Fig. \ref{fig:architecture}f). We include this additional element because of the observation that, while the reconstruction task is useful for pretraining, it may not be optimal for virtual screening. We implement the projector as a 2-layer MLP with hidden dimension $0.5 \cdot D_{embed}$, where $D_{embed}$ is the encoder embedding dimension.

\subsection{Learning Objectives}\label{subsec: learning_objectives}

\textsc{JASPR} is trained with a unified masked data modeling objective \cite{wang2023image}, using a multimodal (HE2ST) objective and optionally including unimodal objectives (e.g., HE2HE, ST2ST). During training, we randomly mask a fraction of the input data with mask ratio $mr$, resulting in $m$ masked tokens, and the decoder learns to reconstruct the masked input \cite{wang2023image}. 
Because of the multimodal nature of the model, there are key design decisions to be made concerning mask tokens and switching experts. We propose that there should be modality specific mask tokens. Furthermore, we investigate two manners in which these mask tokens should be used for the cross-modal scenario HE to ST. For context, we start by describing the unimodal setting.

In \textbf{HE2HE} mode, the input consists of $n \times n$ HE tokens to which we add fixed sin-cos positional embeddings. This sequence is then masked, and encoded with shared MHSA blocks and HE-specific FFNs. For decoding, we reinsert learnable HE mask tokens at the masked positions, add positional embeddings, and predict HE tokens via an HE-specific linear head (HE-Lin). The $\mathcal{L}_{\mathrm{he2he}}$ loss is the reconstruction MSE loss on masked HE tokens. The \textbf{ST2ST} mode follows exactly analogously but with ST-mask tokens and using the ST-specific architecture blocks (Fig. \ref{fig:architecture}b and c).

In \textbf{HE2ST} mode (Fig \ref{fig:architecture}d), the input is encoded in the same way as for HE2HE. However, now we add ST-specific mask tokens to the encoded sequence. This way, the decoder is ``primed'' to recognize it needs to decode differently than in HE2HE mode. The sequence goes through the shared MHSA blocks and HE-FFN layers of the decoder and is transformed with the ST-specific linear head (ST-Lin) resulting in the predicted ST output. The $\mathcal{L}_{\mathrm{he2st}}$ loss is calculated as a supervised MSE loss on the masked tokens. 

Since the HE2ST mode introduces a mismatch in input/output types for HE-FFN in the decoder and ST-Lin, we introduced the joint mode (Fig. \ref{fig:architecture}e). The HE encoding is still the same as in HE2HE, with HE-specific mask tokens. Now, we also append $n^2$ ST-specific mask tokens to the decoder input. Then, we jointly route both sequences through the shared MHSA of the decoder, and split the output to modality-specific FFNs and Lins. The $\mathcal{L}_{\mathrm{joint}}$ loss is a reconstruction MSE loss on the masked tokens for the HE2HE part, and a supervised MSE loss on all $n^2$ predicted ST locations. 

As is standard, there is no direct supervisory signal for the CLS token in the reconstruction task. Therefore, we assess the benefit of adding a symmetric CLIP-style contrastive loss for aligning HE-derived CLS tokens with gene-expression embeddings in the shared latent space (Fig. \ref{fig:architecture}g). Gene expressions within each spatial window are first projected into the latent space via the encoder's ST embedding module and then averaged, mimicking the global aggregation captured by the HE CLS token. We use a symmetric InfoNCE loss in both image-to-gene and gene-to-image directions, computed from pairwise cosine similarities, and define $\mathcal{L}_{\mathrm{con}}$ as the average of the two.

We experiment training \textsc{JASPR} with a combination of the aforementioned modes, evaluating the following loss function as training signal: 

\begin{equation}
\begin{aligned}
\mathcal{L} &= \frac{1}{\lambda_{\mathrm{joint}} + \lambda_{\mathrm{he}} + \lambda_{\mathrm{st}} + 1} \Big[\lambda_{\mathrm{joint}} \mathcal{L}_{\mathrm{joint}} + \lambda_{\mathrm{st}}\mathcal{L}_{\mathrm{st2st}}\\ &
+ (1-\lambda_{\mathrm{joint}})(\mathcal{L}_{\mathrm{he2st}} + \lambda_{\mathrm{he}}\mathcal{L}_{\mathrm{he2he}}) \Big] + \lambda_{\textrm{con}} \mathcal{L}_{\textrm{con}}
\end{aligned}
\label{eq:loss_functions}
\end{equation}
where the possible values of $\{\lambda_{\mathrm{joint}}, \lambda_{\mathrm{he}}, \lambda_{\mathrm{st}}\}$ are $\{0, 0, 0\}$, $\{0, 1, 0\}$, $\{0, 1, 1\}$, $\{1, 0, 0\}$, and $\{1, 0, 1\}$, named HE2ST, HE2HE-sep, ST2ST-sep, HE2HE-joint, and ST2ST-joint, respectively. The value of $\lambda_{\mathrm{con}}$ can be 0 or 1, independently of the other parameters.
After each training epoch, we freeze the encoder and we train the projector for the fully supervised virtual screening task with MSE loss (no masking of HE tokens). 

\subsection{Downstream Application}
We consider several applications of a pretrained \textsc{JASPR} model. Spatial gene expression is predicted from frozen, encoded HE tokens with the ST projector. The frozen encoded HE tokens or CLS tokens are used for whole slide level tasks such as survival prediction by training an aggregator (e.g., mean pooler or ABMIL) \cite{ilse2018attention} (Suppl. Fig. \ref*{fig:downstream_prediction}a). \textsc{JASPR} can also be fine-tuned, though this is not evaluated in our work. 

\section{Experiments}
We performed two sets of experiments: (1) ablative analyses over various \textsc{JASPR} design decisions and (2) domain-specific clinical evaluation tasks. 

\subsection{Training Data}
For training \textsc{JASPR}, we used human breast cancer samples from the HEST-1k \cite{jaume2024hest} dataset. In total, seven patient samples were available with matched fresh-frozen (FF) HE images and Visium gene expression data, totaling 24,697 spots. We reserved one sample as a held-out test set (4,777 spots) and one sample as a validation set (3,539 spots). 

We followed typical preprocessing approaches for both modalities. Specifically, we preprocessed the Visium data by total-count normalization to correct for differences in sequencing depth and performed a log1p-transformation to mitigate potential bias towards large expression values. Finally, we applied a z-score normalization to standardize the gene expression. We used all gene expression values during pretraining and did not restrict to a small subset of highly variable genes. We preprocessed the HE image data by creating tiles centered on the spots at a resolution of 224 $\times$ 224 pixels at 20$\times$ magnification (0.5 microns per pixel). We selected Virchow2~\cite{zimmermann2024virchow2}, based on its superior generalizability~\cite{bareja2025evaluating}, as the histopathology foundation model for tile embedding.

\subsection{Ablation Experiments}
We conducted hyperparameter ablations to estimate the marginal effect of each design choice for \textsc{JASPR} by varying one parameter at a time while holding all others fixed. We performed these on the virtual screening task, and results were aggregated across multiple model sizes (Table \ref{tab:jasper_encoder_decoder_dims}) and across different degrees of multimodal training, using the $\lambda$ settings described above (Eq.~\ref{eq:loss_functions}).
We then selected optimal hyperparameters as those that yielded the largest improvement in performance on the virtual screening task.

\subsection{Clinical Task Experiments} We hypothesized disease specific survival (DSS) and binary estrogen receptor (ER) status prediction, two clinically relevant downstream tasks, may benefit from \textsc{JASPR} pretraining. We evaluated these tasks on FF HE images of human breast cancer samples from The Cancer Genome Atlas dataset (TCGA-BRCA~\cite{weinstein2013cancer}). To avoid patient-level duplication, we retained only one HE image per patient. Labels for DSS can be accessed on the cBioPortal~\cite{cerami2012cbio,gao2013integrative,de2023analysis}, and ER labels were taken from \cite{thennavan2021molecular}. In total, DSS data was available for 985 patients (train: 708, val: 178, test: 99) and ER data was available for 971 patients (train: 698, val: 175, test: 98).

\subsection{Baselines}

We compared frozen \textsc{JASPR}-encoded embeddings to frozen Virchow2 feature embeddings (``baseline''), and to Virchow2 features that were aligned with an ST-derived CLIP-style contrastive objective (``baseline-CLIP'', Suppl. Fig. \ref*{fig:downstream_prediction}b,c). The latter reflects a standard contrastive alignment HE-ST framework \cite{chen2025visual, gindra2025large} where neighborhood structure is not incorporated. By comparing against this setting, we tested whether \textsc{JASPR} contributes value beyond multimodal pretraining alone (i.e., whether its spatially contextual, mask-and-reconstruct formulation contributes additional value).

\subsection{Training Details}
\textsc{JASPR} models were trained using the AdamW optimizer with a batch size of 64 and a learning rate of $2 \times 10^{-4}$ for 400 epochs, with early stopping with a patience of 20 epochs. After each training epoch, the projector was trained for 50 epochs (keeping the encoder frozen) using the same learning rate. We save two versions for each \textsc{JASPR} model, one based on lowest decoder-based validation loss as defined in Eq. \ref{eq:loss_functions}, and one at lowest projector-based virtual screening error. For the survival task, we used the AdamW optimizer with learning rate $1 \times 10^{-5}$, trained for 200 epochs with a patience of 100 epochs. We used batch sizes of 725 for mean aggregation and 256 for ABMIL (considering GPU memory constraints). We used the Cox loss \cite{steyaert2023multimodal} for the survival task. For estrogen receptor status prediction, we used the same optimizer and batch sizes, but with a larger learning rate of $1 \times 10^{-3}$, trained for 500 epochs with a patience of 50 epochs. As a loss function for the ER status task, we used binary cross-entropy. We trained all models with five random initializations and retained the best-performing run across these repetitions. 

\subsection{Evaluation Metrics}
For virtual screening, the Pearson correlation coefficient (PCC) is the most commonly used evaluation metric, as also adopted in the HEST-Benchmark \cite{jaume2024hest} and other related works \cite{jia2023thitogene}. It captures how well the model preserves the relative ranking and co-variation of gene expression across samples. We calculate the PCC on gene-level across spots, and report the mean value across genes. We also report mean squared error (MSE) and mean absolute percentage error (MAPE) in the Supplementary Materials.

For survival prediction, we evaluate patient-level performance using the concordance index (CI), which measures how well predicted risk scores correctly rank patients by observed survival time across all admissible pairs (pairs where the earlier event is uncensored). CI ranges from 0.5 (random) to 1.0 (perfect). For ER subtyping, we evaluate performance using the area under the receiver operator curve (AUC), a typical classification performance metric.

\section{Results}
We evaluated a broad range of preprocessing choices, hyperparameters, training strategies, and model sizes for \textsc{JASPR}. Below, we first present ablation studies on hyperparameters for \textsc{JASPR} for virtual screening. We then evaluate degrees of multimodal training (Eq. \ref{eq:loss_functions}) and model sizes (Suppl. Tables \ref*{tab:jasper_encoder_decoder_dims}, \ref*{tab:jasper_versions}) for virtual screening as well as on a range of clinical and biological applications. 

\begin{figure*}[t]
    \centering
    \includegraphics[width=0.95\textwidth]{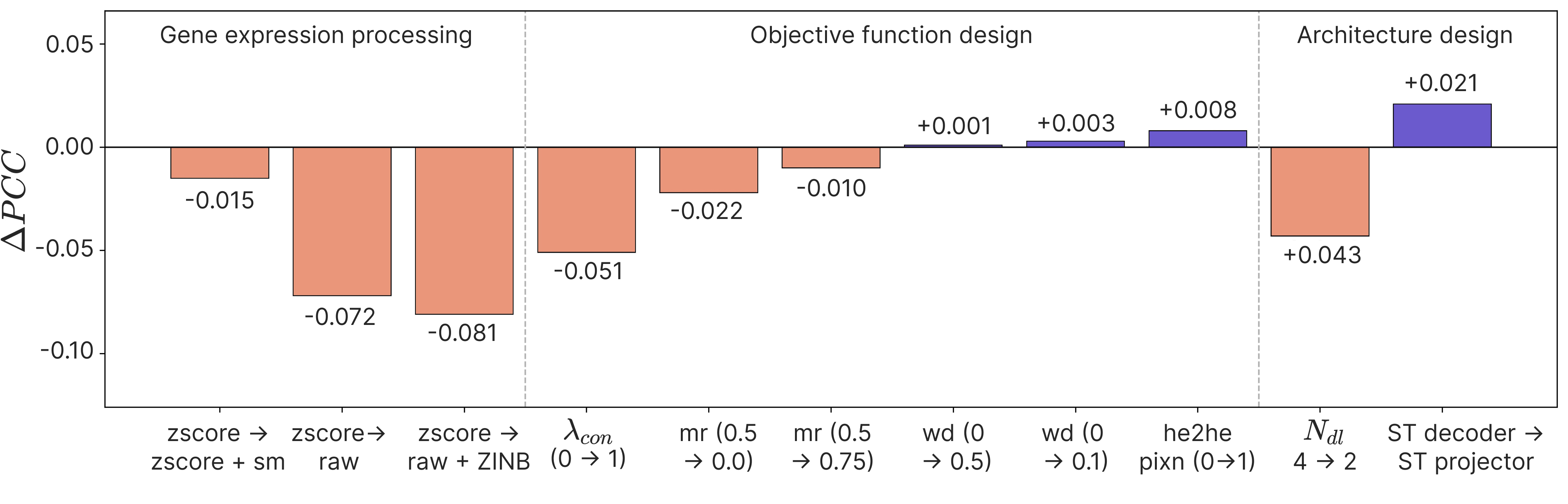}
    \caption{\textsc{JASPR} Virtual screening performance when varying hyperparameters, as specified on the x-axis, averaging over other hyperparameters. sm: smooth, raw: non-zscored, zinb: zero-inflated negative binomial, mr: mask ratio, wd: weight decay, $N_{dl}$: num decoder layers.}
    \label{fig:delta_pcc_barplot2}
\end{figure*}

\subsection{Hyperparameter Ablations}
We consider three categories of algorithmic design choices\textemdash gene expression processing, objective function design, and architecture design\textemdash for the ablative analysis comparing them to variants without these additional elements (Fig.~\ref{fig:delta_pcc_barplot2}). Because gene expression data is sparse and often contains drop-out, we investigated the utility of z-scoring, spatial smoothing as suggested by Shulman et al.~\cite{shulman2024ai}, and a zero-inflated negative binomial (ZINB) loss instead of MSE. Overall, z-scoring without spatial smoothing and using MSE resulted in the best performance and was used for all experiments moving forward.

Next we considered the level of masking and investigated masking ratios of 0.0, 0.5, and 0.75. A masking ratio of 0.5 led to substantially better results than full supervision ($mr=0.0$ leading to a decrease in PCC of 0.022), suggesting that partial masking provides a useful inductive bias, likely by encouraging contextual reasoning and reducing overfitting. A higher masking ratio of 0.75 diminished performance (decrease of 0.010), implying that overly aggressive masking removes too much information and limits effective learning (Fig. \ref{fig:delta_pcc_barplot2}). We observed no benefit from augmenting the objective with a CLIP-based contrastive loss (Fig. \ref{fig:architecture}, Eq. \ref{eq:loss_functions}) nor from normalizing target values when incorporating HE2HE training \cite{he2022masked}. Weight decay provided a small but consistent improvement.

To choose the number of encoder and decoder layers, we followed standard MAE design choices, where the encoder typically has 12/24/32 layers and the decoder 4-8 layers. Given our limited data regime, we focused on the smaller end of this spectrum, fixing the encoder to 12 layers and evaluated decoder depths of four and two layers. We observed that \textsc{JASPR} variants with only two decoder layers achieved substantially lower PCC (Fig. \ref{fig:delta_pcc_barplot2}), indicating insufficient decoder capacity during training. Based on this outcome, we retained the 4-layer decoder configuration for subsequent experiments. For virtual screening, we found that the ST projector outperformed decoder-based prediction (Fig. \ref{fig:architecture} and Fig. \ref{fig:delta_pcc_barplot2}), confirming our intuition that a dedicated projector is needed in addition to the decoder, whose primary role is reconstructing masked tokens during pretraining.

\begin{figure*}[t]
    \centering
    \includegraphics[width=0.95\textwidth]{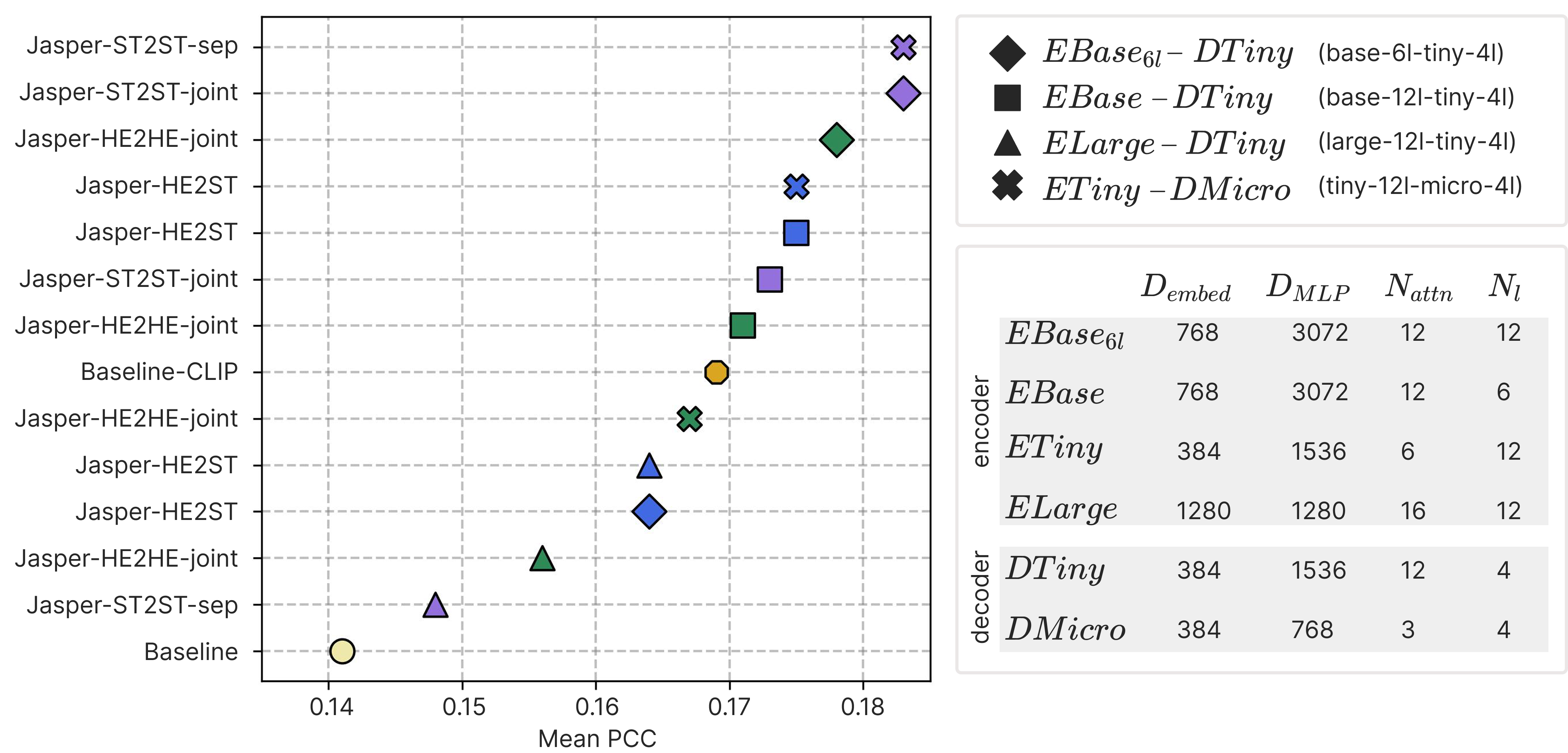}
    \caption{Tile-level Virchow2 baseline versus \textsc{JASPR} performance for virtual screening across all 9,248 genes, in the held-out HEST1k test set. The "-joint" vs. "sep" suffix indicate $\mathcal{L}_{\mathrm{joint}}$ vs. $\mathcal{L}_{\mathrm{he2st}}+\mathcal{L}_{\mathrm{he2he}}$. Model dimensions: hidden ($D_{embed}$) and MLP ($D_{MLP}$) dimensions, $N_{attn}$: num attention heads, $N_l$: num layers.}
\label{fig:st_res_model_sizes}
\end{figure*}

\subsection{Spatial Transcriptomics Prediction Performance}

We next evaluated \textsc{JASPR} for virtual screening across multiple model sizes and increasing degrees of multimodal training (Suppl. Fig. \ref*{fig:downstream_prediction}a, Eq. \ref{eq:loss_functions}). For the settings including the optional unimodal objectives, we compared the -joint and -sep settings and present only the best performing variant.
Here, we present mean PCC across 9,248 predicted genes; additional evaluations (MSE, MAPE, and metric distributions) are provided in Suppl. Tab.~\ref*{tab:st_performance} and Suppl. Fig.~\ref*{fig:scatterplots}-\ref*{fig:histograms}.

Across 9,248 predicted genes, all \textsc{JASPR} variants achieved a higher mean PCC than the tile-level Virchow2 baseline (PCC = 0.141; Fig. \ref{fig:st_res_model_sizes}, Suppl. Table \ref*{tab:st_performance}). Adding a contrastive CLIP-alignment to frozen Virchow2 features increased the baseline PCC by 0.028 (to 0.169), indicating that multimodal contrastive alignment improves the feature space for virtual screening. However, many \textsc{JASPR} models still outperformed this CLIP baseline, suggesting that explicitly modeling spatial context and jointly learning from multimodal information over local neighborhoods captures additional signal beyond tile-level contrastive alignment alone. Smaller \textsc{JASPR} architectures generally performed best, while larger variants (large-12l-small-4l) more often underperformed, likely due to limited amount of training data. The large model size was therefore not used for clinical and biological applications.

Regarding multimodal training objectives, all joint HE2ST objectives outperformed their separate counterparts for the HE2HE setting, showing a clear benefit of keeping the ST-Lin module dedicated to ST inputs and preserving consistent mask-token semantics across the two modalities. When adding ST2ST, the advantage of joint optimization is less pronounced. In this ``most multimodal'' regime, ST-Lin receives direct, modality-matched supervision through the ST2ST objective, which reduces the reliance on joint optimization to anchor ST-Lin behavior. Consequently, separate optimization can become competitive or even advantageous.

Overall, the most multimodal objective (ST2ST) achieved the best mean performance (PCC = 0.183), reached by both base-6l-small-4l-joint and small-12l-micro-4l-sep. The best model without ST2ST (HE2HE) was base-6l-small-4l-joint (PCC = 0.178), underscoring that incorporating ST2ST provides performance gains. Notably, the third-best configuration used HE2ST only (PCC = 0.175), further highlighting that increasingly multimodal objectives (HE2HE and ST2ST) consistently improve performance. 

\subsection{Clinical and Biological Applications}
\begin{figure*}[!h]
\centering

\begin{subfigure}{\textwidth}
  \centering
  \includegraphics[width=0.92\textwidth]{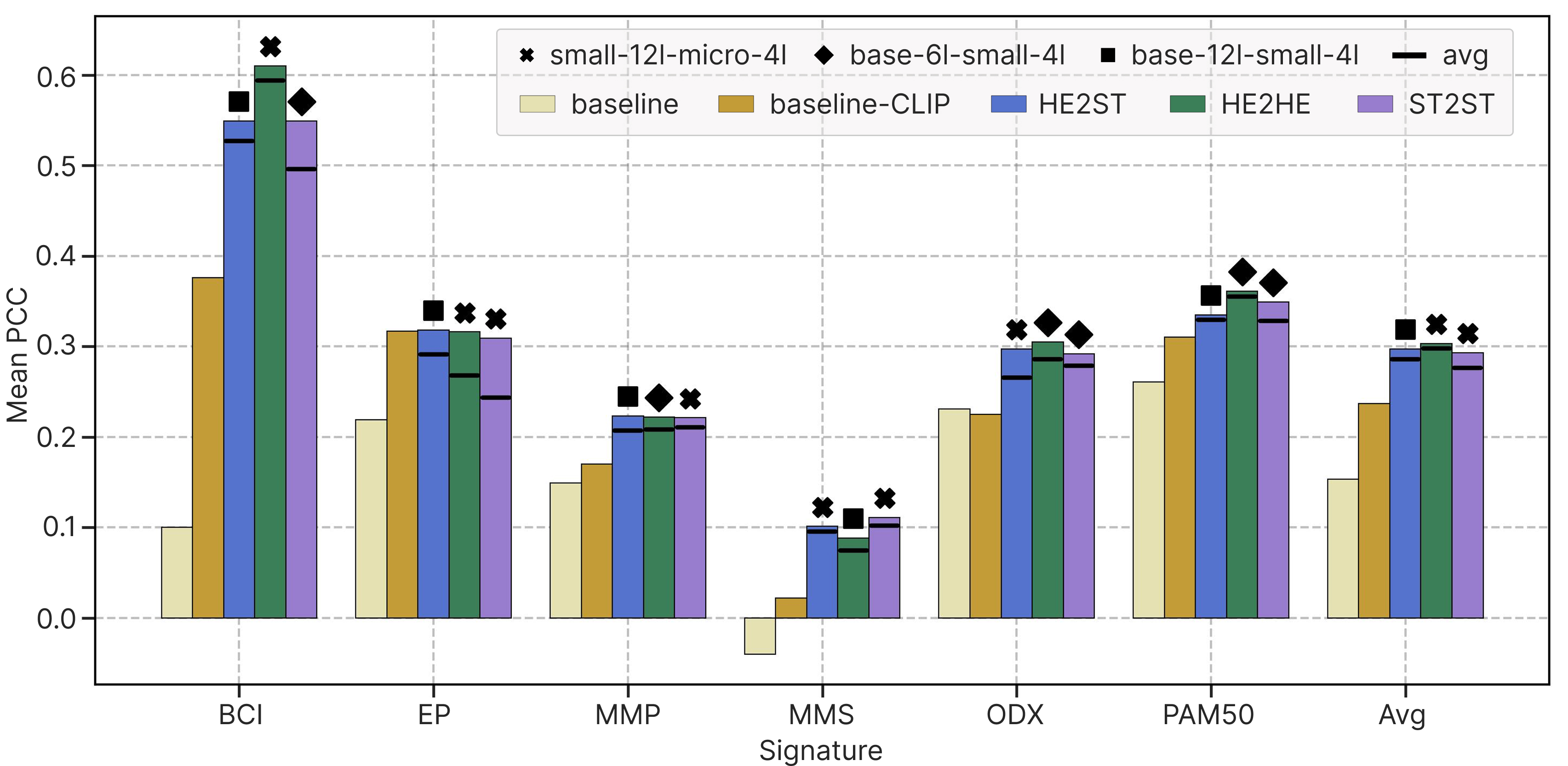}
  \caption{Mean PCC on breast cancer signature gene sets for HEST1k held-out test. BCI: Breast Cancer Index, EP: EndoPredict, MMP: Mammaprint, MMS: Mammostrat, ODX: OncotypeDX, PAM50: Prediction Analysis of Microarray 50, Avg.: Average across signatures. For each model type and signature, the bar represents the highest mean PCC across model sizes. The best size is annotated with an icon on top. Black lines within bars indicate average performance across model sizes. }
  \label{fig:brca_sets}
\end{subfigure}

\par\medskip  

\begin{subfigure}{\textwidth}
  \centering
  \includegraphics[width=\textwidth]{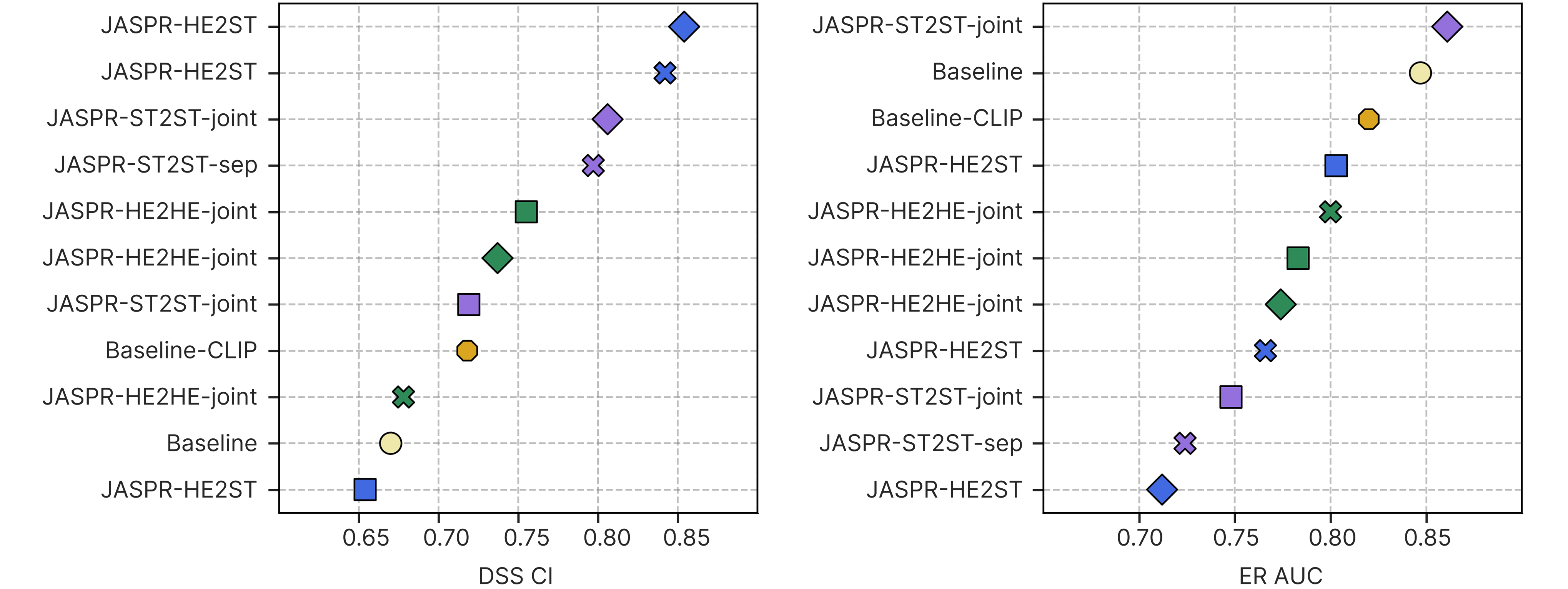}
  \caption{DSS (CI, left) and ER status (AUC, right) prediction on TCGA held-out test set.}
  \label{fig:sub2}
\end{subfigure}

\caption{Clinical and biological applications of \textsc{JASPR} and baselines.}
\label{fig:downstream_evals}
\end{figure*}

As a more clinically relevant measure of virtual screening performance, we predict prognostic breast cancer signatures, which use subsets of the predicted gene expression, without further finetuning (Fig. \ref{fig:downstream_evals}a, Suppl. Table \ref*{tab:signatures}), \textsc{JASPR} consistently improved Pearson correlation (PCC) over both baselines. Averaged across signatures, mean PCC rose from 0.153 (baseline) and 0.237 (baseline-CLIP) to 0.254-0.303 across \textsc{JASPR} configurations. Across \textsc{JASPR} configurations, differences between training objectives and model sizes were small relative to the gap separating \textsc{JASPR} from the baselines (except for EP), with a slight overall advantage for HE2HE.

The performance across baselines and \textsc{JASPR}, as well as the magnitude of improvement by \textsc{JASPR}, varied substantially across signatures. This variation indicates that both the baseline predictability of gene expression from HE and the benefit from HE-ST contrastive alignment and spatial context modeling depend on the underlying biological functions represented in each signature. The strongest gains and best overall performance were observed for the Breast Cancer Index (BCI; mean PCC 0.61 with HE2HE), whereas improvements were considerably smaller for EndoPredict (EP) and PAM50. Notably, the BCI signature is dominated by genes marking tumor cell proliferation and mitotic/chromosomal-instability programs, thereby reflecting the cycling epithelial fraction and linking closely to tumor aggressiveness and recurrence risk. In contrast, while EP and PAM50 also include proliferation markers, they additionally capture estrogen-response and luminal differentiation pathways. EP further incorporates stromal/vascular and inflammatory microenvironment components, whereas PAM50 is primarily oriented toward intrinsic subtype biology (luminal, basal-like, and HER2-enriched). These observations suggest that \textsc{JASPR}'s predictions are most accurate for proliferation-related genes, which exhibit morphologically distinct features visible in HE, whereas genes reflecting hormone signaling, stromal composition, or immune infiltration remain more challenging to predict.

For disease specific survival (DSS) and estrogen receptor (ER) status prediction, we compared mean and ABMIL aggregation. For \textsc{JASPR}, we additionally compared HE-encoded versus encoded CLS tokens (Suppl. Fig. \ref*{fig:downstream_prediction}). Across models and tasks, mean aggregation outperformed ABMIL (DSS CI +0.05 and ER AUC +0.053) and HE-encoded tokens outperformed CLS tokens (DSS CI +0.089 and ER AUC +0.026) (Suppl. Table \ref*{tab:clinical_table}). We hence retained HE-encoded tokens and mean aggregation for further analysis.

For DSS, the baseline achieved a CI of 0.670, while baseline-CLIP improved this to 0.718, indicating that contrastive HE-ST alignment enhances HE-derived feature extraction for DSS. Several \textsc{JASPR} variants further exceeded the CLIP baseline (CI 0.755-0.854), suggesting that our multimodal objectives and the incorporation of spatial context provide additional benefit for DSS prediction (Fig.~\ref{fig:downstream_evals}b). 

For ER status prediction, the baseline achieved an AUC of 0.847. Incorporating the baseline-CLIP embeddings reduced performance to an AUC of 0.820, suggesting that, in this setting, HE-ST alignment does not transfer well to ER-related morphology and may even introduce a negative bias. Similarly, for all \textsc{JASPR} models except one (ST2ST, AUC 0.861), we see a drop in ER status performance with \textsc{JASPR} embeddings, confirming that there is no clear benefit of HE-ST alignment for this task. These observations are consistent with trends observed on breast cancer signature gene sets, again suggesting that \textsc{JASPR}'s HE-ST spatial alignment provides the greatest benefit for cell-cycle and proliferation-related features, which are closely tied to tumor aggressiveness.

\section{Conclusion}

In this work, we introduce \textsc{JASPR}, a joint spatial representation learning framework that addresses a key challenge in computational pathology: integrating histology images (HE) and spatial transcriptomics (ST) data while explicitly modeling spatial context within and across modalities. Our approach employs an encoder-decoder architecture trained with masked autoencoding (MAE) objectives on both unimodal and multimodal inputs. Critically, \textsc{JASPR} combines shared components that capture cross-modal spatial properties with modality-specific experts that preserve the unique characteristics of each data type, enabling the model to learn both joint and modality-specific representations. We performed an extensive sweep of pretraining configurations across multiple model sizes and increasing degrees of multimodality, allowing us to evaluate the contributions of HE2HE and ST2ST pretraining beyond HE2ST modeling.

For virtual ST prediction across 9,248 genes in breast cancer, all \textsc{JASPR} variants outperformed the HE-only tile-level baseline, and many also surpassed a tile-level baseline based on contrastive HE-ST embeddings, demonstrating clear advantages of incorporating spatial context and jointly learning from multimodal information. Performance generally increased with multimodality: the best results were obtained with the most multimodal objective (ST2ST), followed by HE2HE and then HE2ST alone.

\textsc{JASPR} also markedly outperformed both tile-level baselines on prognostically relevant gene subsets in breast cancer, with the most pronounced improvements for the Breast Cancer Index, a gene set predominantly reflecting cell-cycle and proliferation processes, and that strongly correlates with tumor aggressiveness and recurrence risk \cite{bartlett2024validation, jilderda2025validation}. Consistent with these gains, \textsc{JASPR} also showed clear value over the two baselines for disease specific survival prediction. For processes related to hormone signaling and stromal/immune features, explicit HE-ST alignment did not show measurable benefits. For estrogen receptor status prediction, HE-ST alignment could even degrade performance, including for the baseline and many \textsc{JASPR} variations, suggesting these phenotypes may be sufficiently captured by histology alone or are not helped by cross-modal alignment under the current setup. 

Overall, the most multimodal \textsc{JASPR} variant (ST2ST, base-6l-small-4l-joint) consistently achieved strong performance across all evaluated tasks, underscoring that joint modeling of spatial HE-ST context learns improved, transferable representations between tissue morphology and gene expression beyond what unimodal pretraining or tile-level HE-ST alignment can capture.

\par\vfill\par

\clearpage  

\bibliographystyle{splncs04}
\bibliography{main}

\clearpage
\setcounter{page}{1}
\setcounter{section}{0}
\setcounter{subsection}{0}
\setcounter{table}{0}
\setcounter{figure}{0}
\maketitlesupplementary

\vspace{1cm}

\section{Supplementary Figures}

\begin{strip}
    \centering
    \includegraphics[width=\textwidth]{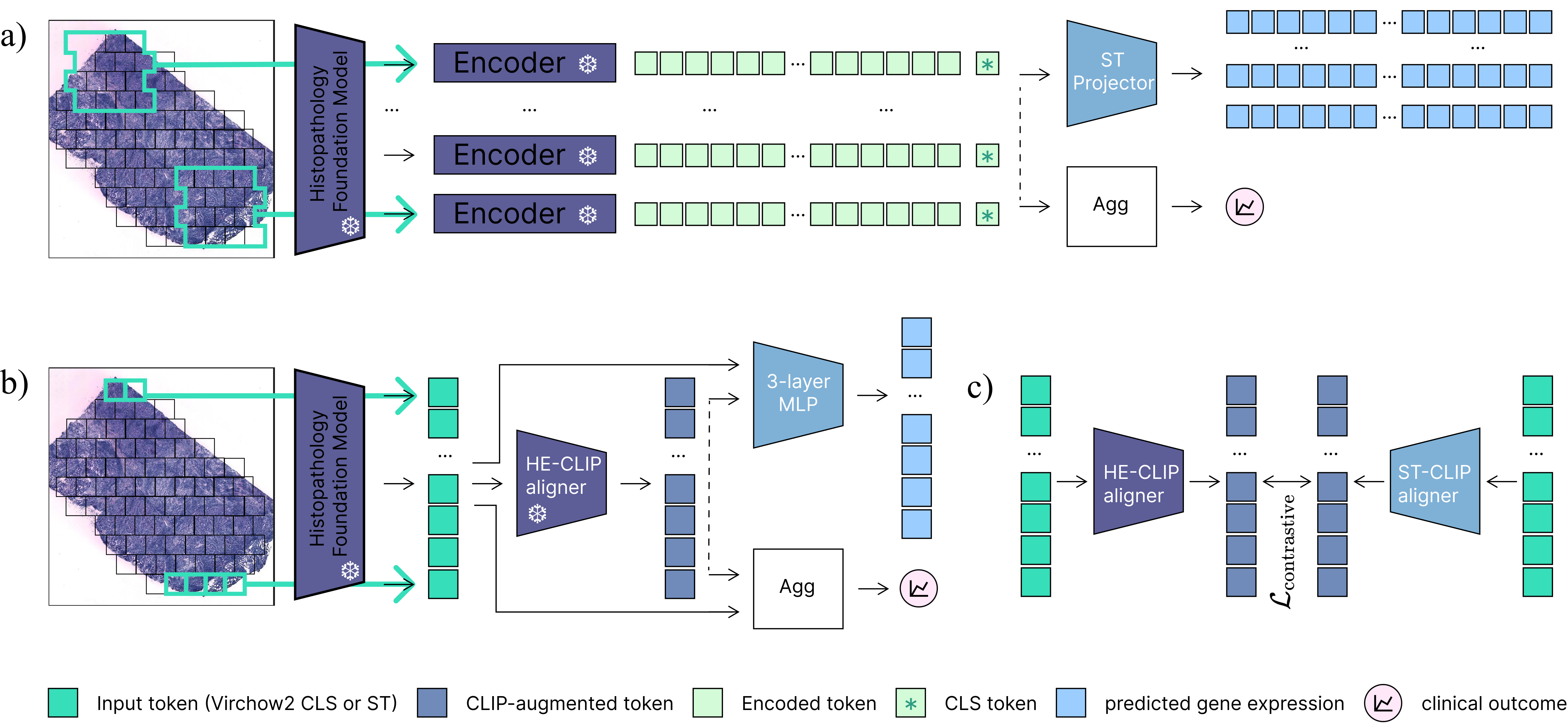}
     \captionof{figure}{Downstream application of \textsc{JASPR} features and compared baseline architectures. a) \textsc{JASPR}. b) Baseline. c) CLIP contrastive alignment training.}
    \label{fig:downstream_prediction}
\end{strip}

\begin{figure*}
    \centering
    \includegraphics[width=\textwidth]{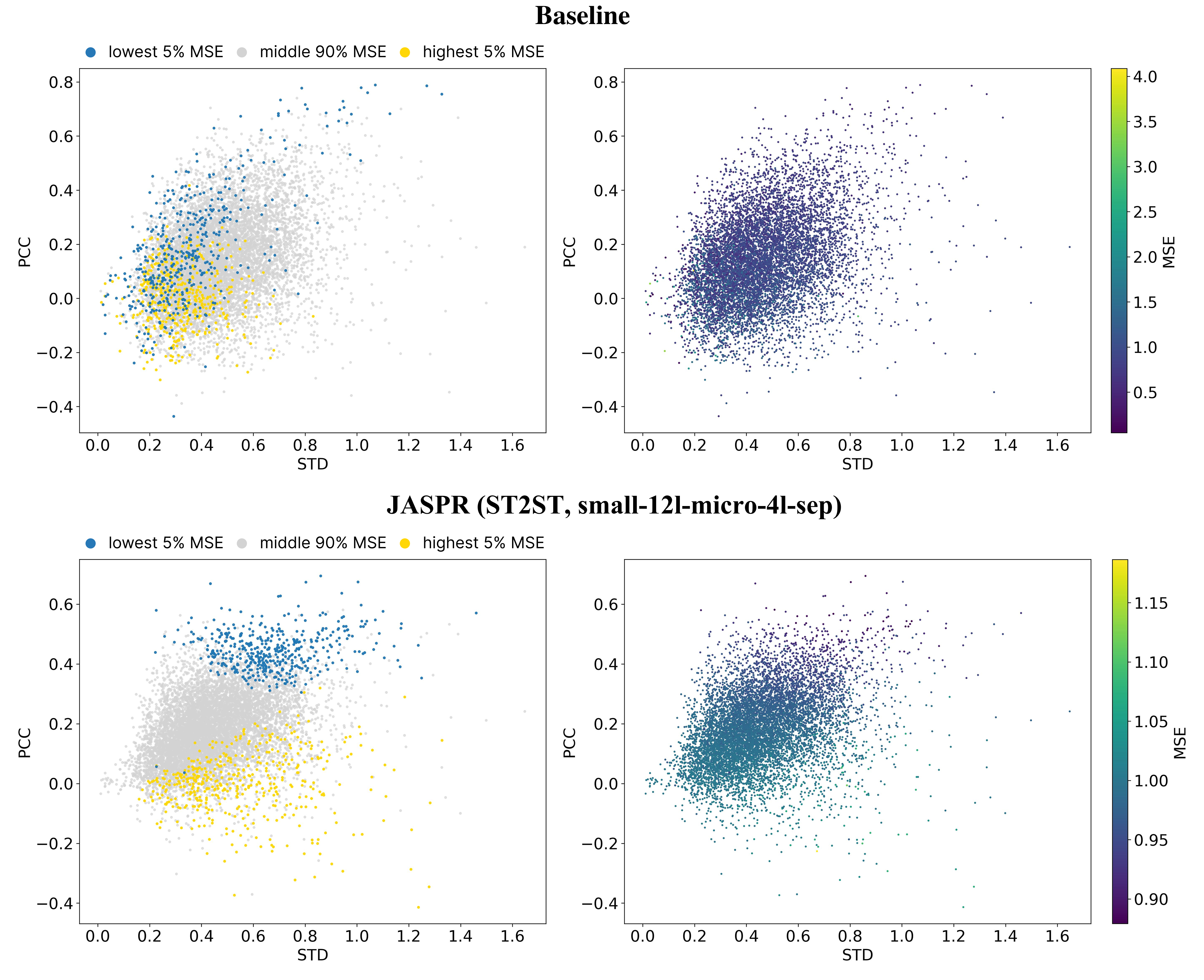}
    \caption{Gene-level Pearson Correlation Coefficient (PCC) versus ground-truth, non-zscored standard deviation (STD), colored by Mean Squared Error (MSE).}
    \label{fig:scatterplots}
\end{figure*}

\begin{figure*}
    \centering
    \includegraphics[width=\textwidth]{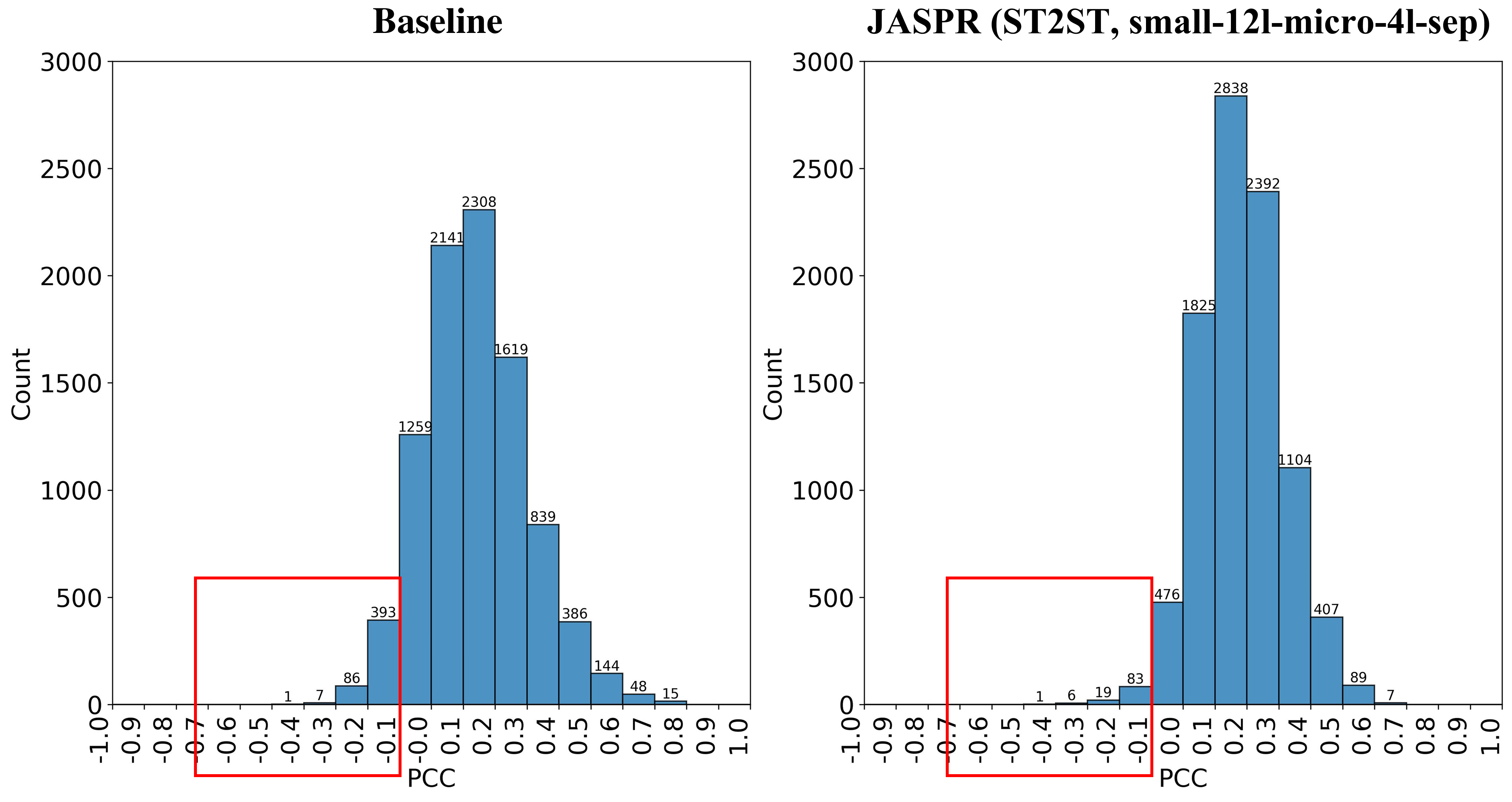}
    \caption{Histogram of gene-level Pearson Correlation Coefficient (PCC).}
    \label{fig:histograms}
\end{figure*}

\clearpage

\section{Supplementary Tables}

\renewcommand{\arraystretch}{1.5}

\noindent
\makebox[\textwidth]{%
\begin{minipage}{\textwidth}
\centering

\captionsetup{
    width=\textwidth,
    justification=centering,
    singlelinecheck=false
}
\captionof{table}{Overview of methods that model histology images and paired genomics data.}
\label{tab:sota_overview}
    \centering
    \begin{NoHyper}
    \begin{tabular}{p{5cm}p{2cm}p{2cm}p{2cm}p{1.5cm}}
    \toprule
    Existing methods & Genomics \newline technology & Spatial ST \newline context & Spatial HE \newline context & Joint \newline training \\
    \midrule
    Patch-based HE and \newline bulk genomics \cite{kather2020pan, pizurica2023whole, schaumberg2016h, chen2020classification, qu2021genetic, wang2021predicting} & Bulk DNA, \newline bulk RNA & \xmark & \xmark & \xmark\\
    Slide-based HE \newline and bulk genomics \cite{alsaafin2023learning, schmauch2020deep, pizurica2024digital} & Bulk RNA & \xmark & \cmark & \xmark\\
    Patch-based HE and \newline spatial  genomics \cite{shulman2024ai, he2020integrating} & Visium & \xmark & \xmark & \xmark\\
    Slide-based HE and \newline spatial genomics \cite{nonchev2025deepspot, pang2021leveraging, jia2023thitogene} & Visium & \xmark & \cmark & \xmark \\
    Contrastive slide-HE and \newline bulk genomics \cite{jaume2024transcriptomics} & Bulk RNA & \xmark & \cmark & \cmark \\
    Contrastive patch-HE \newline and  spatial genomics \cite{chen2025visual, redekop2025spade, gindra2025large, xie2023spatially} & Xenium \newline Visium & \xmark & \xmark & \cmark \\
    Joint slide-image* and \newline spatial genomics \cite{kong2025spatia} & Xenium & \xmark  & \cmark {\tiny *microscopy images, not HE} & \cmark \\
    \midrule
    \textbf{\textsc{JASPR} (ours)} & Visium & \cmark & \cmark & \cmark \\
    \bottomrule
    \end{tabular}
    \end{NoHyper}
\end{minipage}
}

\renewcommand{\arraystretch}{1}

\begin{table*}
\centering
\begin{minipage}{0.45\textwidth}
\centering
\caption{Model dimensions. $D_{embed}$: hidden dimension, $D_{MLP}$: dimension of MLP, $N_{attn}$: number of attention heads, $N_l$: number of layers.}
    \label{tab:jasper_encoder_decoder_dims}
    \centering
    \begin{tabular}{lccccc}
        \toprule
        name & $D_{embed} $ & $D_{MLP}$ & $N_{attn}$ & $N_l$  \\
        \midrule
        \multicolumn{5}{c}{$Encoder$} \\
        \noalign{\vskip 5pt}
        \hdashline
        \noalign{\vskip 5pt}
        $ESmall$ & 384 & 1536 & 6 & 12 \\
        $EBase_{6l}$ & 768 & 3072 & 12 & 6 \\
        $EBase$ & 768 & 3072 & 12 & 12 \\
        $ELarge$ & 1280 & 1280 & 16 & 12 \\
        &&&&\\
        \multicolumn{5}{c}{$Decoder$} \\
        \noalign{\vskip 5pt}
        \hdashline
        \noalign{\vskip 5pt}
        $DSmall$ & 384 & 1536 & 12 & 4 \\
        $DSmall_{2l}$ & 384 & 1536 & 12 & 2 \\
        $DMicro$ & 192 & 768 & 3 & 4 \\
         \bottomrule
    \end{tabular}
\end{minipage}\hfill
\begin{minipage}{0.45\textwidth}
\centering
\caption{Encoder-decoder combinations for \textsc{JASPR}.}
    \label{tab:jasper_versions}
    \centering
    \begin{tabular}{lll}
        \toprule
        $Encoder$ & $Decoder$ & name \\
        \midrule
        $EBase$ & $DSmall$ & base-12l-small-4l \\
        $EBase_{6l}$ & $DSmall$ & base-6l-small-4l\\
        $EBase_{6l}$ & $DSmall_{2l}$ & base-6l-small-2l \\
        $ESmall$ & $DMicro$ & small-12l-micro-4l \\
        $ELarge$ & $DSmall$ & large-12l-small-4l \\
         \bottomrule
    \end{tabular}
\end{minipage}
\end{table*}

\begin{table*}
    %\footnotesize
    \centering
    \caption{Performance of baseline versus \textsc{JASPR} models across all 9,248 genes, measured by Mean Absolute Percentage Error (MAPE), Mean Squared Error (MSE) and mean Pearson Correlation Coefficient (PCC). Entries that outperform baseline are shaded in green. Best performance within grouped model type is indicated in bold. Best performance per metric overall across all models is underlined. See Suppl. Note 1 for discussion.} % 
    \label{tab:st_performance}
    %\small
    \begin{tabular}{p{3.5cm}p{3.5cm}p{1.5cm}p{1.5cm}p{1.5cm}}
    \toprule
 & architecture & MAPE & MSE & PCC \\
    \midrule
baseline & Virchow2 + MLP & \textbf{1.170} &  \underline{\textbf{0.933}} & \textbf{0.141} \\
\noalign{\vskip 5pt}
\hdashline
\noalign{\vskip 5pt}
baseline contrastive  & $\mathrm{Virchow2}_{\mathrm{CLIP}}$ + MLP & \textbf{1.170} & \textbf{0.937} & \cellcolor{PineGreen!20}\textbf{0.169} \\
\noalign{\vskip 5pt}
\hdashline
\noalign{\vskip 5pt}
& \textsc{JASPR} & & & \\
\noalign{\vskip 5pt}
\hdashline
\noalign{\vskip 5pt}
 HE2ST & base-12l-small-4l & \cellcolor{PineGreen!20}{\textbf{1.049}} & 0.984 & \cellcolor{PineGreen!20}{\textbf{0.175}} \\
 & base-6l-small-4l & \cellcolor{PineGreen!20}{1.102} & 0.985 & \cellcolor{PineGreen!20}{0.164} \\
 & large-12l-small-4l & \cellcolor{PineGreen!20}{1.090} & \textbf{0.983} & \cellcolor{PineGreen!20}{0.164} \\
 & small-12l-micro-4l & \cellcolor{PineGreen!20}{1.086} & 0.988 & \cellcolor{PineGreen!20}{\textbf{0.175}} \\
 &&&& \\
HE2HE & base-12l-small-4l-joint & \cellcolor{PineGreen!20}{1.144} & 0.977 & \cellcolor{PineGreen!20}{0.171} \\
 & base-6l-small-4l-joint & \cellcolor{PineGreen!20}{\textbf{1.094}} & \textbf{0.973} & \cellcolor{PineGreen!20}{\textbf{0.178}} \\
 & large-12l-small-4l-joint & \cellcolor{PineGreen!20}{\textbf{1.094}} & 0.979 & \cellcolor{PineGreen!20}{0.156} \\
 & small-12l-micro-4l-joint & \cellcolor{PineGreen!20}{1.119} & 0.981 & \cellcolor{PineGreen!20}{0.167} \\
 &&&& \\
ST2ST  & base-12l-small-4l-joint & \cellcolor{PineGreen!20}{1.068} & \textbf{0.978} & \cellcolor{PineGreen!20}{0.173} \\
 & base-6l-small-4l-joint & \cellcolor{PineGreen!20}{1.097} & 0.982 & \cellcolor{PineGreen!20}{\textbf{\underline{0.183}}} \\
 & large-12l-small-4l-sep & \cellcolor{PineGreen!20}{\textbf{\underline{1.048}}} & 0.989 & \cellcolor{PineGreen!20}{0.148} \\
 & small-12l-micro-4l-sep & \cellcolor{PineGreen!20}1.057 & 0.983 & \cellcolor{PineGreen!20}{\textbf{\underline{0.183}}} \\
 \bottomrule
\end{tabular}
\end{table*}

\begin{table*}[]
\caption{Performance of baseline versus \textsc{JASPR} models on subsets of genes measured by mean Pearson Correlation Coefficient. BCI: Breast Cancer Index, EP: EndoPredict, MMP: Mammaprint, MMS: Mammostrat, ODX: OncotypeDX, PAM50: Prediction Analysis of Microarray 50, Avg.: Average across signatures. Entries that outperform baseline are shaded in green. Best performance within grouped model type is indicated in bold. Best performance per metric overall across all models is underlined.} 
    \label{tab:signatures}
    \centering
    \begin{tabular}{p{1.4cm}p{3.2cm}p{1cm}p{1cm}p{1cm}p{1cm}p{1cm}p{1cm}p{1cm}}

    \toprule
    &  & BCI & EP & MMP & MMS & ODX & PAM50 & Avg. \\
    \midrule
& baseline & 0.100 & \textbf{0.219} & 0.149 & \textbf{-0.040} & \textbf{0.231} & 0.261 & \textbf{0.153} \\

\noalign{\vskip 5pt}
\hdashline
\noalign{\vskip 5pt}

& baseline contrastive & \cellcolor{PineGreen!20}{\textbf{0.376}} & \cellcolor{PineGreen!20}{\textbf{0.317}} & \cellcolor{PineGreen!20}{\textbf{0.170}} & \cellcolor{PineGreen!20}{\textbf{0.022}} & \textbf{0.225} & \cellcolor{PineGreen!20}{\textbf{0.310}} & \cellcolor{PineGreen!20}{\textbf{0.237}} \\

\noalign{\vskip 5pt}
\hdashline
\noalign{\vskip 5pt}
 & \textsc{JASPR} & & & \\
\noalign{\vskip 5pt}
\hdashline
\noalign{\vskip 5pt}

HE2ST  & base-12l-small-4l & \cellcolor{PineGreen!20}{\textbf{0.549}} & \cellcolor{PineGreen!20}{\textbf{\underline{0.318}}} & \cellcolor{PineGreen!20}{\textbf{\underline{0.223}}} & \cellcolor{PineGreen!20}{0.096} & \cellcolor{PineGreen!20}{0.261} & \cellcolor{PineGreen!20}{\textbf{0.335}} & \cellcolor{PineGreen!20}{\textbf{0.297}} \\
 & base-6l-small-4l & \cellcolor{PineGreen!20}{0.541} & \cellcolor{PineGreen!20}{0.286} & \cellcolor{PineGreen!20}{0.201} & \cellcolor{PineGreen!20}{0.088} & \cellcolor{PineGreen!20}{0.238} & \cellcolor{PineGreen!20}{0.322} & \cellcolor{PineGreen!20}{0.279} \\
 & small-12l-micro-4l & \cellcolor{PineGreen!20}{0.492} & \cellcolor{PineGreen!20}{0.269} & \cellcolor{PineGreen!20}{0.196} & \cellcolor{PineGreen!20}{\textbf{0.101}} & \cellcolor{PineGreen!20}{\textbf{0.297}} & \cellcolor{PineGreen!20}{0.331} & \cellcolor{PineGreen!20}{0.281} \\
 &&&&&&& \\
 HE2HE & base-12l-small-4l-joint & \cellcolor{PineGreen!20}{0.574} & \cellcolor{PineGreen!20}{0.222} & \cellcolor{PineGreen!20}{0.197} & \cellcolor{PineGreen!20}{0.088} & \cellcolor{PineGreen!20}{0.294} & \cellcolor{PineGreen!20}{0.354} & \cellcolor{PineGreen!20}{0.288} \\
 & base-6l-small-4l-joint & \cellcolor{PineGreen!20}{0.597} & \cellcolor{PineGreen!20}{0.265} & \cellcolor{PineGreen!20}{\textbf{0.222}} & \cellcolor{PineGreen!20}{0.060} & \cellcolor{PineGreen!20}{\textbf{\underline{0.305}}} & \cellcolor{PineGreen!20}{\textbf{\underline{0.361}}} & \cellcolor{PineGreen!20}{0.302} \\
 & small-12l-micro-4l-joint & \cellcolor{PineGreen!20}{\textbf{\underline{0.610}}} & \cellcolor{PineGreen!20}{\textbf{0.316}} & \cellcolor{PineGreen!20}{0.206} & \cellcolor{PineGreen!20}{0.075} & \cellcolor{PineGreen!20}{0.259} & \cellcolor{PineGreen!20}{0.351} & \cellcolor{PineGreen!20}{\textbf{\underline{0.303}}} \\
  &&&&&&& \\
ST2ST & base-12l-small-4l-joint & \cellcolor{PineGreen!20}{0.423} & \cellcolor{PineGreen!20}{0.228} & \cellcolor{PineGreen!20}{0.202} & \cellcolor{PineGreen!20}{0.098} & \cellcolor{PineGreen!20}{0.272} & \cellcolor{PineGreen!20}{0.303} & \cellcolor{PineGreen!20}{0.254} \\
 & base-6l-small-4l-joint & \cellcolor{PineGreen!20}{\textbf{0.549}} & 0.194 & \cellcolor{PineGreen!20}{0.208} & \cellcolor{PineGreen!20}{0.096} & \cellcolor{PineGreen!20}{\textbf{0.292}} & \cellcolor{PineGreen!20}{\textbf{0.349}} & \cellcolor{PineGreen!20}{0.281} \\
 & small-12l-micro-4l-sep & \cellcolor{PineGreen!20}{0.516} & \cellcolor{PineGreen!20}{\textbf{0.309}} & \cellcolor{PineGreen!20}{\textbf{0.221}} & \cellcolor{PineGreen!20}{\textbf{0.111}} & \cellcolor{PineGreen!20}{0.271} & \cellcolor{PineGreen!20}{0.332} & \cellcolor{PineGreen!20}{\textbf{0.293}} \\
 \bottomrule
\end{tabular}
\end{table*}

\begin{table*}[]
\caption{Baseline versus \textsc{JASPR} performance on downstream tasks: Disease Specific Survival (DSS) and estrogen receptor (ER) status measured by AUC. }
    \centering
\begin{tabular}{p{1.5cm}p{3.5cm}p{1.5cm}p{1.5cm}p{1.5cm}p{1.5cm}p{1.5cm}}
\toprule
  &  & input  & agg. & DSS CI & BRCA ER \\
 \midrule
 \multicolumn{2}{c}{baseline} & &  abmil & 0.667 & \textbf{0.855}  \\
\multicolumn{2}{c}{baseline} & & mean & \textbf{0.670} & 0.847 \\
 \noalign{\vskip 5pt}
\hdashline
\noalign{\vskip 5pt}
\multicolumn{2}{c}{baseline contrastive} & & abmil & \cellcolor{PineGreen!20}{\textbf{0.725}} & 0.846 \\
\multicolumn{2}{c}{baseline contrastive} & & mean & \cellcolor{PineGreen!20}{0.718} & 0.820 \\
 \noalign{\vskip 5pt}
\hdashline
\noalign{\vskip 5pt}
HE2ST & base-12l-small-4l & cls & abmil & 0.627 & 0.775 \\
 &  & cls & mean & 0.552 & 0.755 \\
 &  & patch & abmil & \cellcolor{PineGreen!20}{0.716} &  0.735 \\
 &  & patch & mean & 0.654 & \textbf{0.803} \\
 & base-6l-small-4l & cls & abmil & 0.493 & 0.701 \\
 &  & cls & mean & \cellcolor{PineGreen!20}{0.803} &  0.755 \\
 &  & patch & abmil & \cellcolor{PineGreen!20}{0.702} & 0.768 \\
 &  & patch & mean & \cellcolor{PineGreen!20}{\textbf{0.854}} & 0.712 \\
 & small-12l-micro-4l & cls & abmil & 0.564 & 0.614 \\
 &  & cls & mean & 0.588 & 0.717 \\
 &  & patch & abmil & \cellcolor{PineGreen!20}{0.716} & 0.716 \\
 &  & patch & mean & \cellcolor{PineGreen!20}{0.842} & 0.766 \\
&&&&& \\
HE2HE & base-12l-small-4l-joint & cls & abmil & 0.657 & 0.703 \\
 &  & cls & mean & \cellcolor{PineGreen!20}{0.696} & 0.747 \\
 &  & patch & abmil & \cellcolor{PineGreen!20}{0.740} & 0.687 \\
 &  & patch & mean & \cellcolor{PineGreen!20}{\textbf{0.755}} & 0.783 \\
 & base-6l-small-4l-joint & cls & abmil & \cellcolor{PineGreen!20}{0.684} &  0.633 \\
 &  & cls & mean & 0.624 & \textbf{0.800} \\
 &  & patch & abmil & 0.642 & 0.780 \\
 &  & patch & mean & \cellcolor{PineGreen!20}{0.737} & 0.774 \\
 & small-12l-micro-4l-joint & cls & abmil & \cellcolor{PineGreen!20}{0.713} &  0.766 \\
 & & cls & mean & 0.615 & 0.770 \\
 & & patch & abmil & 0.585 & 0.696 \\
 & & patch & mean & \cellcolor{PineGreen!20}{0.678} & \textbf{0.800} \\
 &&&& \\
ST2ST & base-12l-small-4l-joint & cls & abmil & 0.576 & 0.600 \\
 &  & cls & mean & \cellcolor{PineGreen!20}{0.713} & 0.731 \\
 &  & patch & abmil & \cellcolor{PineGreen!20}{\textbf{\underline{0.869}}}  & 0.671 \\
 &  & patch & mean & \cellcolor{PineGreen!20}{0.719} & 0.748 \\
 & base-6l-small-4l-joint & cls & abmil & 0.618 &  0.719 \\
 &  & cls & mean & 0.663 & 0.800 \\
 &  & patch & abmil & \cellcolor{PineGreen!20}{0.672} &  0.778 \\
 &  & patch & mean & \cellcolor{PineGreen!20}{0.806} & \cellcolor{PineGreen!20}{\textbf{\underline{0.861}}} \\
 & small-12l-micro-4l-sep & cls & abmil & 0.648 & 0.752 \\
 &  & cls & mean & 0.663 & 0.798 \\
 &  & patch & abmil & 0.621 & 0.805 \\
 &  & patch & mean & \cellcolor{PineGreen!20}{0.797} & 0.724 \\
 \bottomrule
\end{tabular}
    
    \label{tab:clinical_table}
\end{table*}

\clearpage

\section{Supplementary Notes}

\textbf{Suppl. Note 1: Spatial Transcriptomics prediction performance}

To evaluate ST prediction performance, we focused on mean PCC performance in the main text, as this is the standard evaluation metric in the field \cite{jaume2024hest}. This is largely because many downstream analyses emphasize relative expression patterns (e.g., differential expression and gene ranking) rather than exact agreement in absolute expression scale.

Here, we go into more detail, also covering MSE and MAPE, and metric distributions, to more thoroughly assess for which genes \textsc{JASPR} and the tile-level baseline model reach good performance.

MSE measures the average squared difference between predicted and true expression values and is therefore sensitive to large errors. However, MSE does not account for gene-specific variance. Since some genes exhibit very low variance, a model can achieve a low average MSE by performing well on less informative genes. MAPE complements this by expressing the absolute error relative to the true value, thereby placing greater emphasis on genes with higher variability. Because MAPE can be disproportionately inflated by genes with very small expression values (as in our data), we computed MAPE per gene and report the median across genes in Suppl. Table \ref{tab:st_performance} for robustness. The relative ordering of models is unchanged when using the mean MAPE; however, the mean is dominated by these extreme cases and the reported values inflate to $\sim 10^9$. As is standard, MSE and PCC are reported as mean across genes.

To examine why the baseline attains the lowest mean MSE while underperforming on PCC and MAPE, we further analyzed gene-level behavior. We examined scatter plots of per-gene PCC versus per-gene standard deviation, with points colored by gene-level MSE. This reveals a different error profile between models (Suppl. Fig. \ref{fig:scatterplots}). 

For \textsc{JASPR}-based models, PCC increases more systematically with gene standard deviation, suggesting that the learned representations better track expression variation in higher-variance (and typically more informative) genes. In contrast, the baseline shows a large cluster of genes with very low MSE that also have very low standard deviation and poor PCC, suggesting a prediction of near-constant expression for low-variance genes, an effect that reduces average MSE without preserving gene-wise variation. At the same time, we observe a smaller set of high-variance genes for which the baseline also attains low MSE, indicating that the baseline performs well at the extremes: it fits many low-variance genes and a handful of very high-variance ones. Conversely, \textsc{JASPR} models capture many genes with high standard deviation and high PCC - but at higher MSE than the baseline, implying that they recover the correct relative trends (high correlation) while being miscalibrated in absolute magnitude for some genes. 

To more clearly quantify how many genes fall into different PCC ranges, we include a histogram of predicted values in Suppl. Fig. \ref{fig:histograms}. 
The baseline model predicts a large number of genes with negative PCC, whereas negative PCC values are much rarer for \textsc{JASPR}. While the baseline captures some  genes in the highest PCC bins, \textsc{JASPR} predicts a larger overall number of genes with good PCC ($\mathrm{PCC}>0.2$).

In conclusion, the baseline seems to perform particularly well on a subset of genes with either very low or very high standard deviation for which it reaches low MSE and low PCC, while \textsc{JASPR} does better across the board.

\clearpage

\end{document}